\documentclass[letterpaper, 10 pt, conference]{ieeeconf}  

\IEEEoverridecommandlockouts                         

\usepackage{xcolor}
\usepackage{graphicx}
\usepackage{amsmath}
\usepackage{booktabs}
\usepackage{algorithm}
\usepackage{algorithmic}
\usepackage{amssymb}
\usepackage{multirow}
\usepackage{cite}
\usepackage{tablefootnote}
\usepackage{amsmath}
\usepackage{subfigure}
\usepackage{amsfonts}
\usepackage{wrapfig}
\usepackage{bm}

\title{\LARGE \bf
Dynamic Scenario Representation Learning for Motion Forecasting with Heterogeneous Graph Convolutional Recurrent Networks 
}

\author{Xing Gao, Xiaogang Jia, Yikang Li, and Hongkai Xiong
\thanks{Xing~Gao, Xiaogang~Jia, and Yikang~Li are with the Shanghai AI Lab, Shanghai 200232, China. (\textit{Corresponding author: Xing Gao, E-mail:    gxyssy@163.com.)}}%
\thanks{H. Xiong is with the Department of Electronic Engineering, Shanghai Jiao Tong University, Shanghai 200240, China.}%
}

\begin{document}
\maketitle
\thispagestyle{empty}
\pagestyle{empty}

\begin{abstract}
Due to the complex and changing interactions in dynamic scenarios, motion forecasting is a challenging problem in autonomous driving. Most existing works exploit static road graphs to characterize scenarios and are limited in modeling evolving spatio-temporal dependencies in dynamic scenarios. In this paper, we resort to dynamic heterogeneous graphs to model the scenario. Various scenario components including vehicles (agents) and lanes,  multi-type interactions, and their changes over time are jointly encoded. Furthermore, we design a novel heterogeneous graph convolutional recurrent network, aggregating diverse interaction information and capturing their evolution, to learn to exploit intrinsic spatio-temporal dependencies in dynamic graphs and obtain effective representations of  dynamic scenarios. Finally, with a motion forecasting decoder, our model predicts realistic and multi-modal future trajectories of agents and outperforms  state-of-the-art published works on several motion forecasting benchmarks.
\end{abstract}

\section{Introduction}
Motion forecasting that predicts future trajectories of surrounding agents is an important and core module in autonomous driving systems for a safe and comfortable self-driving. 
Prediction is inherently uncertain and multi-modal.
For example, a car may follow a vehicle ahead or change lanes on a congested road.
Fortunately, historical trajectories of agents and high definition (HD) maps provide  cues to perceive context information and render predictions feasible. 

Exploitation of such information is nontrivial, however, because of 
(i) highly heterogeneous scenario components \cite{varadarajan2021multipath}, including surrounding agents, lanes, traffic lights, \textit{etc.}; (ii) complex and multiple interactions, like agent-agent interaction and agent-road interaction; (iii) interlaced spatio-temporal  information, such as the trajectories of agents. 

\begin{figure}[tp]
\vskip 0.04in
\begin{center}
\centerline{\includegraphics[width=0.98\columnwidth]{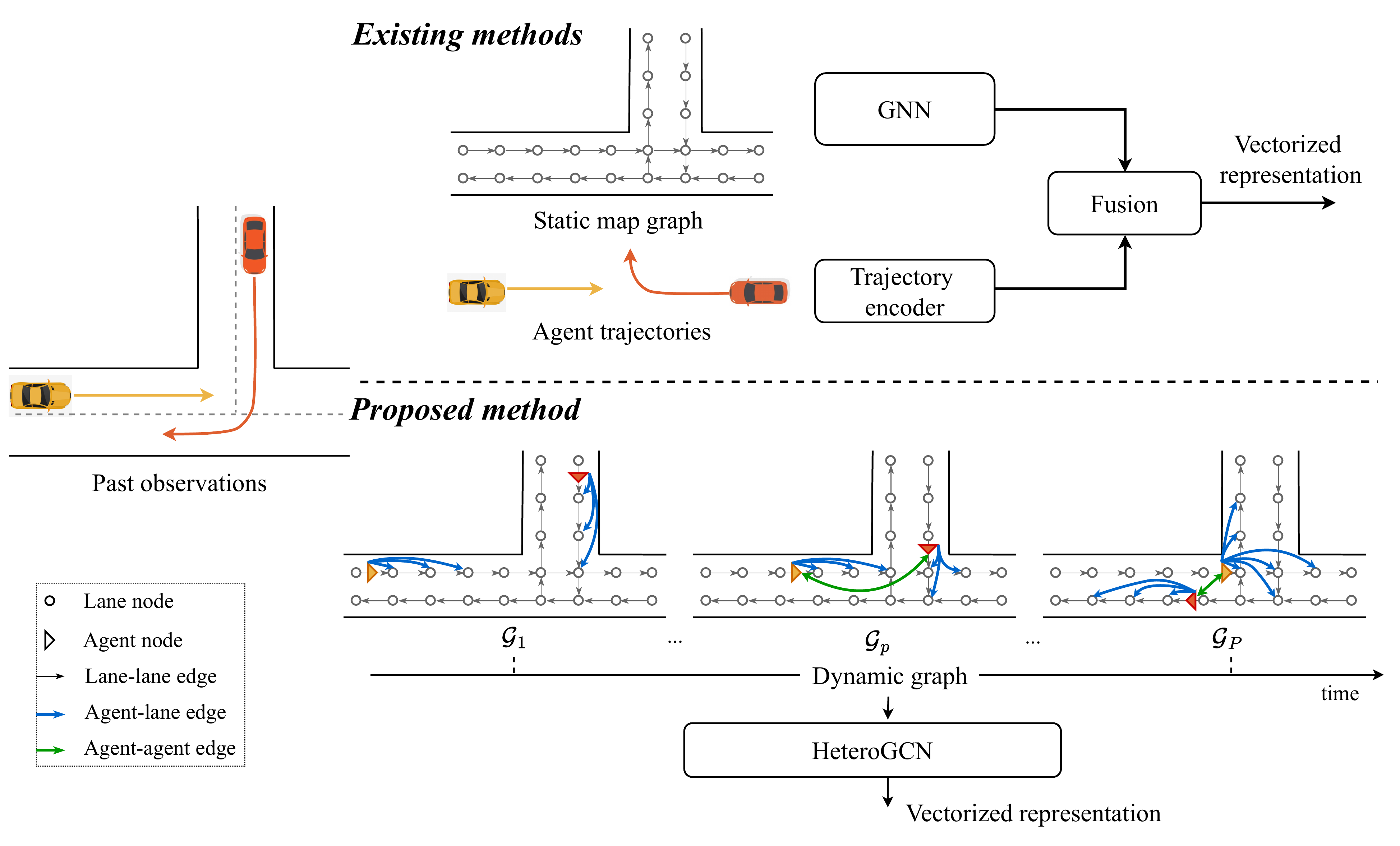}}
\caption{A high-level demonstration of differences in dynamic scenario modeling and spatio-temporal information processing between existing methods and the proposed strategy.}\label{fig.0}
\end{center}
\vskip -0.2in
\end{figure}

Several methods \cite{chai2019multipath,cui2019multimodal,bansal2018chauffeurnet,casas2018intentnet,hong2019rules} rasterize the scenario to encode road information and historical trajectories of agents and exploit convolutional neural networks (CNNs) to learn complex interactions in the scenario. However, the sparse and irregular topology of road networks is inefficient to capture with rasterization-based representations  and long-range interactions are difficult to process with local kernels of CNNs.  Alternatively, a variety of graph-based methods \cite{gao2020vectornet,liang2020learning,zeng2021lanercnn,gilles2021gohome,mo2022multi} have been  proposed recently. They typically employ graphs to represent irregular road networks and design various graph convolutional networks (GCNs) to further learn representations, such as VectorNet \cite{gao2020vectornet}, LaneGCN \cite{liang2020learning}, and LaneRCNN \cite{zeng2021lanercnn}.

However, these methods still have limitations in the following two aspects, as illustrated in Fig.~\ref{fig.0}. (i) Dynamic scenario modeling. Existing graph-based methods employ \textit{static graphs} to model either road networks \cite{liang2020learning,zeng2021lanercnn,gilles2021gohome} or agent-agent interaction networks \cite{mo2022multi}, instead of the whole scenario. These static homogeneous graphs fail to characterize the evolution in dynamic scenarios and 
the diversity of scenario components and their interactions. (ii) Joint spatio-temporal information processing.  With static graph representation, existing methods  typically process temporal decencies and  spatial interactions separately and combine them with late-fusion strategies. This will hinder the model from capturing the intrinsic correlation of spatio-temporal information.

\begin{figure*}[tp]
\begin{center}
\centerline{\includegraphics[width=1.8\columnwidth]{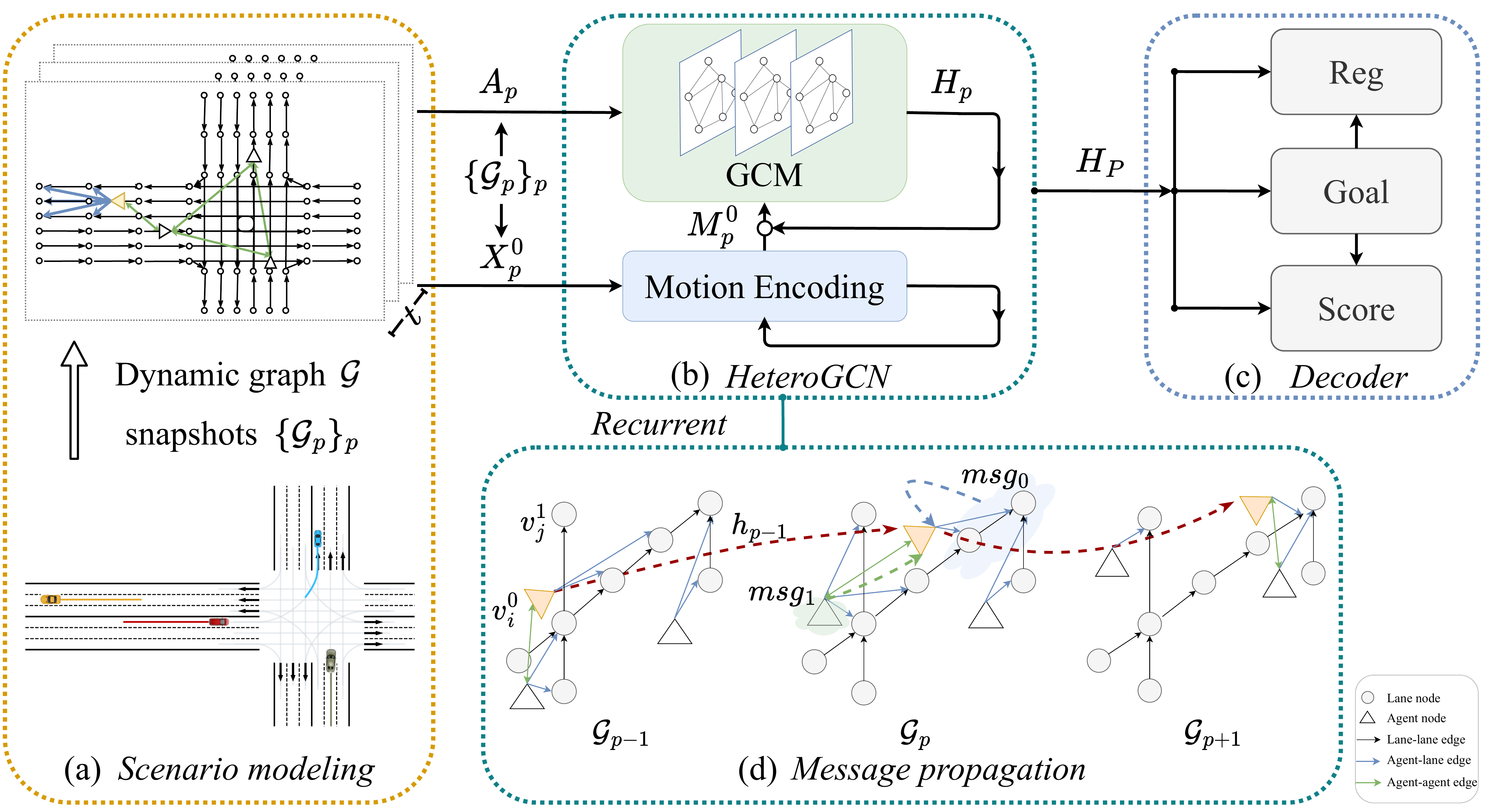}}
\caption{Illustration of the framework. The dynamic graph for scenario modeling is demonstrated with a randomly selected agent (yellow triangle) in (a). Specifically, agent nodes and lane nodes are respectively represented as triangles and circles, and various interactions between them are indicated by edges in different colors. Based on the dynamic graph, the proposed heterogeneous graph convolutional recurrent network further captures high-order information of scenarios and forecasts future trajectories of agents, as shown in (b) and (c). Specially, message passing strategies for dynamic graphs in HeteroGCN are demonstrated in (d), with various colored dashed lines indicating message propagation along different types of edges.} \label{fig.1}
\label{fig:node_spec}
\end{center}
\vskip -0.1 in
\end{figure*}

To tackle these problems, we propose a novel dynamic scenario representation learning method for motion forecasting, called HeteroGCN, as illustrated in Fig.~\ref{fig.1}. 
HeteroGCN first models dynamic scenarios with the help of \textit{dynamic heterogeneous graphs}.
A heterogeneous graph convolutional recurrent network  is further designed, which consists of heterogeneous graph convolution operators and motion encoding modules, for joint spatio-temporal information processing. The benefits
of HeteroGCN are summarized as below.
\begin{itemize}
\item The proposed scenario modeling strategy explicitly encodes dynamic attributes of agents  and multiple time-varying interactions in the scenario into the signal and topology of dynamic graphs, through mining from historical trajectories of agents, and inherits the advantages of static graph based road network encoding methods in capturing its sparse and irregular structure.
\item The proposed graph convolutional recurrent networks gradually aggregate spatio-temporal information through processing the evolution of the topology and signal of dynamic graphs, which facilities to explore intrinsic spatio-temporal dependencies.
\item The designed heterogeneous graph convolution operator handles multiple types of nodes and interactions jointly yet differently, which permits to fuse road information and agent features at different time slots directly.
\end{itemize}
Based on the scenario representation, we predict future trajectories of agents with a motion forecasting decoder.
It is shown to outperform state-of-the-art published motion forecasting methods on the challenging large-scale Argoverse and Argoverse2 motion forecasting benchmarks. 

\section{Related Works}
We overview a collection of motion forecasting methods, especially focusing on vectorized (sparse) encoding based scenario representations.

Some early methods \cite{alahi2016social, gupta2018social} exploit LSTMs or GRUs to encode temporal information of agents but ignore spatial interactions between agents and roads. In addition to LSTMs, multi-head attention is exploited in 
Jean \cite{9197340} to handle interactions between agents,  but road map information is still missing.
Recently, TPCN \cite{ye2021tpcn} and DCMS \cite{ye2022dcms} introduce techniques from point cloud processing to learn scenario representation. They employ a spatial module to extract features of waypoints as well as map information and a temporal module to capture sequential information of agents. 
Besides, several transformer based models \cite{ngiam2021scene,girgis2021latent,zhou2022hivt} utilize a stack of self-attention and cross-attention modules to capture complex interactions across agents, road lines, and temporal states, in a decoupled way. Multipath++ \cite{varadarajan2021multipath} designs a multi-context gating,  an efficient variant of cross-attention, to fuse various interactions and further employs ensemble to improve the multimodality of predictions.
 
Alternatively, graph neural networks (GNNs) have recently attended much attention in the field of motion forecasting with graph data capturing the sparse and irregular topology of road networks efficiently. For example, Vectornet \cite{gao2020vectornet} proposes a two-level graph neural network to produce vectorized representation of scenario components. A local graph network extracts features of each component, including trajectories of agents and road polygons, and global interactions between these objects are further processed with another graph convolution layer. Based on the representation from Vectornet, TNT \cite{zhao2020tnt} designs a goal-based prediction decoder, and DensenTNT \cite{gu2021densetnt} further improves the goal prediction module with a dense goal candidate set.
However, the topology of road networks is not effectively exploited in Vectornet with a fully-connected global graph. To address it, LaneGCN \cite{liang2020learning} models road networks with lane graphs and designs a graph convolutional network to capture the complex topology of lane graphs. Similarly,  Gohome \cite{gilles2021gohome} and THOMAS \cite{gilles2022thomas} encode the structure of road networks with the help of lanelet-level graphs and  output a heatmap to predict positions of agents. Furthermore, LaneRCNN \cite{zeng2021lanercnn} proposes an agent-specific subgraph to combine historical motions of agents and their respective local contexts, and handles interactions through pooling in the global lane graph. In addition to lane graphs, HEAT \cite{mo2022multi} employs heterogeneous graphs to model various agent-agent interactions.

In contrast with these methods, we propose a  dynamic heterogeneous graph to represent dynamic scenarios including agents and road networks, instead of a static lane-graph for road networks \cite{liang2020learning,zeng2021lanercnn,gilles2021gohome}. Furthermore,  we design a novel  heterogeneous graph convolutional recurrent network to address the heterogeneity in nodes and edges of a graph and capture its time-varying topology and signal.

\section{Notions and Preliminaries}
\textbf{Notions.} In this paper, we represent a discrete-time  dynamic graph and its discrete snapshots  with $\mathcal{G}$ and $\mathcal{G}_p= \{\mathcal{V}_p, \mathcal{E}_p\}_{p=1}^P$,  and use the subscript $p$ to indicate terms belonging to the $p$-th snapshot $\mathcal{G}_p$. Specifically, $\mathcal{V}_p$ indicates vertex set and $\mathcal{E}_p$ is edge set. The topology of $\mathcal{G}_p$ is represented by adjacency matrix $A_p$, and signals or features on nodes in the graph are denoted as $X_p$, with $\bm{x_{p,i}}=x_p(v_i)$ corresponding to features of node $v_i \in \mathcal{V}_p$ and $X_p= \left[\bm{x_{p,1}}, \bm{x_{p,2}}, \dots, \bm{x_{p,n}} \right]^T$, $n=|\mathcal{V}_p|$. If the graph is further heterogeneous, we use superscripts to distinguish different types. For example, the set of vertices is $\mathcal{V}_p = \bigcup_{z } \mathcal{V}_p^z$ and the set of edges is $\mathcal{E}_p = \bigcup_{r} \mathcal{E}_p^r$, with $\mathcal{V}_p^z$ and $\mathcal{E}_p^r$ indicating different types of node sets and edge sets.  $X_p^z$ as a sub-matrix of $X_p$ indicates features of nodes  $v_i^{z_i} \in \mathcal{V}_p^z$. Besides, matrices and vectors are generally represented by capital letters and bold lowercase letters. 

\subsection{Graph Convolution}
Convolution has been generalized from Euclidean data to graph data to enable deep neural networks to capture the irregular topology of graphs recently. Generally, the definition of graph convolution derives from either spectral graph theory or message passing strategies, corresponding to spectral graph convolutional networks~\cite{bruna2013spectral,defferrard2016convolutional,khasanova2017graph,bianchi2019graph} and spatial graph convolutional networks \cite{gilmer2017neural,velikovi2018graph,hamilton2017inductive,xu2018how}. 
Spatial graph convolutional networks are much prevalent with their flexible and diverse designs of message passing and aggregation schemes. They generally consist of two stages, message passing in neighborhoods and node feature  update. For each node $v_i \in \mathcal{V}$, features on its neighbors $v_j \in N(v_i)$ are first processed with a aggregator function $\mathop{\Box}$ to generate a message $msg(v_i)$, node $v_i$ then updates its features based on the message from its neighbors and its previous layer representation $x_l(v_i)$.
\begin{gather}
    msg(v_i)= \mathop{\Box}\limits_{v_j \in N(v_i)} \varphi \big( x_l(v_j), x_l(v_i)\big)\\
    x_{l+1}(v_i)=\xi \big(x_l(v_i), msg(v_i) \big),
\end{gather}
where $x_l(v_i)$ indicates the feature of node $v_i$ in the $l$-th layer of graph neural networks, $\varphi(\cdot)$ and $\xi(\cdot)$ represent message functions and update functions, respectively.

\section{Dynamic Graphs for Scene Modeling}
\label{sec.graph}
We exploit dynamic graphs to explicitly model driving scenarios and their evolution over time. Specially, all of the scenario components, including agents and roads, and their multi-type interactions are jointly encoded in the topology of graphs.

Given a sequence of historical scenes, we first group them every $\tau$ discrete time, and construct a snapshot for each group instead of per discrete time to reduce computation. In other words, the $p$-th snapshot $\mathcal{G}_p$ is built with information at time $t= \tau*(p-1)-T'+1, \tau*(p-1)-T'+2, \dots,  \tau*p-T'$,  with $T'=\tau  *  P $ and $p=1, 2,\dots,P$. Due to the diversity of scenario components and their interactions, each snapshot is constructed as a \textit{heterogeneous} graph.

Specifically, we consider two kinds of nodes  \textit{i.e.}, agents and lane segments,  and four kinds of edges, including $<{\rm agent}, {\rm agent}>$, $<{\rm agent}, {\rm lane}>$, $<{\rm lane}, { \rm agent}>$, and $< {\rm lane}, {\rm lane}>$, in graphs. Without loss of generality, we assume that the set of vertices $\mathcal{V}_p$  keeps the same in different $\mathcal{G}_p$, except their attributes vary with $p$.  For the sake of clarity, we omit the subscript $p$ of nodes  below,  when without causing confusion.

\textbf{Nodes.} We take each agent in the scenario as an agent node $v^0_i \in \mathcal{V}^0$, and each lane segment as a lane node $v^1_i \in \mathcal{V}^1$. 
Agent node features are designed as  their states 
and state displacements. Specifically, for the $p$-th group of historical scenes, there are $\tau$ discrete locations and displacements per agent node $v_i^0 \in \mathcal{V}_p$,  at time  $t= \tau*(p-1)-T'+1, \tau*(p-1)-T'+2, \dots,  \tau*p-T'$, and we adopt its state at $t=\tau*p-T'$, \emph{i.e.,} $s_{\tau*p-T'}(v_i^0)$, and all the displacements between these $\tau$ adjacent states as agent node features $\bm{x_{p,i}^0}=x_p(v_i^0)$ in $\mathcal{G}_p$, $p=1, 2,\dots,P$.
Similarly, the features of lane nodes are initially adopted from their locations and location displacements, and the features are further processed with a two-layer graph convolutional network,  a variant of GraphSage \cite{hamilton2017inductive}, to encode map topologies.

\textbf{Lane-lane edges} ($\mathcal{E}^0$) are linked following the topology of road networks. Specifically, if  lane segment $v_i^1$ and  segment  $v_j^1$ are directly connected in sequence according to the road direction, a directed edge $e_{i,j}^0= <v_i^1, v_j^1> \ \in \mathcal{E}^0$ is constructed to connect $v_i^1$ to $v_j^1$. Since road topology is static,  $\mathcal{E}^0$ is the same in different $\mathcal{G}_p$ and the subscript $p$ is omitted.

\textbf{Lane-agent edges} ($\mathcal{E}^1$) are built on lane-lane edges $\mathcal{E}^0$. Specially, we encode \textbf{\textit{the speed, heading, and location of agents}} at corresponding time into the topology of $\mathcal{G}_p$. Specifically, for historical scenes at time $t= \tau*(p-1)-T'+1, \tau*(p-1)-T'+2, \dots,  \tau*p-T'$,  we first find the $k$ nearest lane nodes for each agent node $v_i^0$ based on  its coordinate $\bm{c_{p,i}}=c_t(v_i^0)$,  $t=\tau*p-T'$, and discard some of them belonging to opposing lanes.\footnote{On intersections, we keep all $k$ nodes, considering that agents may make a U-turn.} Then, starting from these nodes, we perform a depth-first-search (DFS) along edges in $\mathcal{E}^0$, exploring as far as possible until all nodes reachable from the source node within the maximum depth have been found. The maximum depth of DFS is based on the average speed of the agent, the average gap between adjacent lane nodes, and the time to forecast. Finally, we link the agent node $v_i^0$ to the explored lane nodes $\{ v_j^1\}_j$ with edges $\mathcal{E}^1_p=\{e_{i,j}^1=<v_i^0,v_j^1 > \}$. Meanwhile, we obtain \textbf{agent-lane} edges $\mathcal{E}^2_p= \{e_{j,i}^2=<v_j^1, v_i^0> | \forall e_{i,j}^1 \in \mathcal{E}^1_p \}$.

\textbf{Agent-agent edges} are constructed in accordance with the distance between their locations. Specifically, the distance between $v_i^0$ and  $v_j^0$ is computed with the $\ell_1$ norm to approximate the distance of agents along roads.
\begin{equation}
    d_p(v_i^0, v_j^0)= \|c_p(v_i^0)-c_p(v_j^0)\|_1, \quad p=1,2,\dots, P.
\end{equation}
Then, they are connected with direct edges $e_{i,j}^3, e_{j,i}^3 \in \mathcal{E}_p^3$  in $\mathcal{G}_p$,  if $d_p(v_i^0, v_j^0) < \delta_{aa}$, with $\delta_{aa}$ as a hyperparameter denoting the distance threshold.

\section{Heterogeneous Graph Convolutional Recurrent Networks}
We first outline the framework of the proposed graph neural networks,  then elaborate its components and introduce the learning procedure.
\subsection{Framework}
As presented in Fig.~\ref{fig.1}, the whole framework consists of an encoder to learn vectorized representations of dynamic scenarios and their components  and a decoder to produce future trajectories of agents.

The encoder is designed as a  heterogeneous graph convolutional recurrent network. It takes in dynamic scenarios that are modeled as dynamic graphs and aggregates spatio-temporal information of dynamic graphs with a graph convolution module (GCM) and a motion encoding module.  For each snapshot $\mathcal{G}_p$, $p=1, 2, \dots, P$, the motion encoding module as a causal module produces motion features  $M_p^0$ for each agent node  from its  historical trajectory. The motion features $M_p^0$ together with  spatial features $H_{p-1}^0$ from $\mathcal{G}_{p-1}$ are then processed with a spatio-temporal gate $g(\cdot)$ to update the features of agent nodes at the beginning of the graph convolution module,
\begin{equation}
    H_{p-1}^0 \leftarrow g(H_{p-1}^0, M_{p}^0), \quad p=1, 2, \dots, P,
\end{equation}
with $H_0^0$ initialized as a zero matrix. As an example, $g(\cdot)$ is implemented as summation in this paper.  A stack of heterogeneous graph convolution operators in the GCM further aggregate information to processes complex high-order interactions in  snapshots $G_p$.
\begin{equation}
    H_{p}=GCM(H_{p-1}, A_p),  \quad p=1,2,\dots,P,
\end{equation}
with features of lane nodes initialized as map features $H_0^1=X_1^1$. Notably, the encoder is a recurrent network and the GCM is shared across different snapshots $\{ \mathcal{G}_p\}_{p=1}^P$ of $\mathcal{G}$.

Based on encoded representation $H_P$ from the encoder,  the decoder  composed of three branches further predicts $K$ goals $\{\bm{\hat{s}_{T}}^{k}\}$, future trajectories $\hat{S}_f^{k}= \left[ \bm{\hat{s}_{1}}, \bm{\hat{s}_{2}},...,\bm{\hat{s}_{T-1}} \right]$, $k=1,2,\dots, K$, and their corresponding confidence scores, respectively.

\subsection{Motion Encoding Module}
The motion encoding module aims to extract motion features of agents from their raw trajectory features $\{X_1^0, X_2^0, \dots,X_p^0\}$. Specifically, two 2-layer multilayer perceptrons (MLPs) first map the raw agent features $X_p^0$,  positions and displacements, in each snapshot  $\mathcal{G}_p$ into feature spaces and produce corresponding embeddings. Notably, these MLPs are common to different snapshots. Taking in these two embedding sequences, two 1-layer GRUs further process information across snapshots, respectively.  Finally, a linear layer fuses the features corresponding to positions and displacements from respective GRUs to obtain motion features $M_p^0$, $p=1, 2, \dots, P$.

\subsection{Heterogeneous Graph Convolution Operator}
\label{sec.gcn}
To represent various nodes and capture their multiple relationships, we design a heterogeneous graph convolution operator through message passing. For the sake of clarity, we ignore the subscript $p$ of items in this subsection.

Generally,  we adopt different message passing schemes along distinct types of edges in $\mathcal{G}$. For any node $v_i^{z_i}$, messages $msg_r$ to  $v_i^{z_i}$ are first aggregated  within each of its distinct neighborhoods $\{N^r(v_i^{z_i})\}_r$, $ r=0,1,\dots,3$, and these messages are further combined and processed to produce its context representation $msg(v_i^{z_i}) \in \mathbb{R}^d$, with $d$ indicating the number of node hidden features.
\begin{equation}\label{e.msg}
\begin{split}
msg(v_i^{z_i})= {\rm ReLU} \big( \sum_{r} \max_{v_j \in N^r} msg_r(v_j^{z_j} \rightarrow v_i^{z_i}) \big), \\ \ z_i, z_j \in \{0,1\} \ \text{and} \ r=0,1,\dots,3.
\end{split}
\end{equation}
Along each type of edges, the message from $v_j^{z_j}$ to $v_i^{z_i}$ is defined as a function of node feature $\bm{h_j} \in \mathbb{R}^d$ ($H=\left[\bm{h_{1}}, \dots, \bm{h_j}, \dots, \bm{h_{n}} \right]^T \in \mathbb{R}^{n \times d}$) and its relationship $r(v_i^{z_i},v_j^{z_j}) \in \mathbb{R}^d$ with the target node
\begin{gather}
  msg_r(v_j^{z_j} \rightarrow v_i^{z_i})=  f(\bm{h_j}, r(v_i^{z_i}, v_j^{z_j})) .
\end{gather}
The $f(\cdot): \mathbb{R}^{2d} \rightarrow \mathbb{R}^{d}$ is designed  as an MLP in order to  approximate a suitable function by learning from data.  The relationship $r(v_i^{z_i},v_j^{z_j})$ between nodes is computed based on their feature similarity and coordinate displacement
\begin{equation}
r(v_i^{z_i},v_j^{z_j})=\psi( (Q_z\bm{h_i}) \odot \bm{h_j},\bm{c_i}-\bm{c_j}),
\end{equation}
where $\psi(\cdot)$ represents a learnable non-linear transformation, $Q_z$ is a learnable affine matrix, $\bm{c_i}=c(v_i^{z_i})$ denotes node coordinates, and $\odot$ indicates the Hadamard product. 

Based on the context message $msg(v_i^{z_i})$ and a self-transformation $\nu_z(\cdot)$,
the convolution operator outputs node features
\begin{gather}
\begin{split}
\bm{h_{p,i}}={\rm ReLU} \big(W_{z_i} \cdot [ \nu_{z_i}(\bm{h_{p-1,i}}) \ \| \  msg(v_i^{z_i})] \big),  \\ z_i= 0, 1 \ \text{and} \ p=1,2, \dots, P, \label{e.gcn1}    
\end{split}
\end{gather}
where $W_{z_i} \in \mathbb{R}^{d \times 2d}$ represents a learnable parameter matrix.
Specially, a shortcut is further introduced to facilitate information and gradient propagation, and then Eq.~\eqref{e.gcn1} becomes
\begin{gather}\label{e.gcn}
\begin{split}
\bm{h_{p,i}}={\rm ReLU} \big(W_{z_i} \cdot [ \nu_{z_i}(\bm{h_{p-1,i}}) \ \| \  msg(v_i^{z_i})] + \bm{h_{p-1,i}} \big).
\end{split}
\end{gather}

\begin{table*}[tp]
\vskip 0.1in
\centering
 \caption{Results on the testing set of Argoverse motion forecasting benchmark (* indicating results with ensemble). }\label{t:1}
\setlength{\tabcolsep}{4mm}{
\begin{tabular}{l|ccc|ccc|c}
	\toprule
Method	&\multicolumn{3}{c|}{$K=1$}&\multicolumn{4}{c}{$K=6$}\\
	       & minADE & minFDE& MR & minADE & minFDE& MR& \textbf{B-minFDE} \\
	\midrule
	
	LaneRCNN&1.69&3.69&0.57&0.90&1.45&0.12&2.15\\
        Jean&1.74&4.24&0.69&1.00&1.42&0.13&2.12\\
	Prime&1.91&3.82&0.59&1.22&1.56&0.12&2.10\\
	LaneGCN & 1.71 & 3.78 & 0.59&0.87 &1.36 &0.16&2.05\\
	Gohome&1.69&3.65&0.57&0.94&1.45&0.10&1.98\\
	DenseTNT(minFDE)&1.68 &3.63 &0.58  &0.88 &1.28 &0.13&1.98 \\
	THOMAS&1.67&3.59&0.56&0.94&1.44&0.10&1.97\\
	SceneTransformer&1.81 &4.06 &0.59 &0.80&1.23 &0.13&1.89\\
	Home+Gohome*&1.70&3.68&0.57&0.89&1.29&\textbf{0.09}&1.86\\
Multipath++*&1.62&3.61&0.56&\textbf{0.79}&1.21&0.13&1.79\\
	\hline
	HeteroGCN &1.62&3.52&0.55&0.82&1.19&0.12&1.84\\
	HeteroGCN-en* &\textbf{1.57}&\textbf{3.41}&\textbf{0.54}&\textbf{0.79}&\textbf{1.16}&0.12&\textbf{1.75} \\
	\bottomrule
\end{tabular}}
\vskip 0.1in
\centering
 \caption{Results on the testing set of Argoverse2 motion forecasting benchmark (* indicating results with ensemble). 
 }\label{t.av2}
\setlength{\tabcolsep}{4mm}{
\begin{tabular}{l|ccc|ccc|c}
	\toprule
Method	&\multicolumn{3}{c|}{$K=1$}&\multicolumn{4}{c}{$K=6$}\\
	       & minADE & minFDE& MR & minADE & minFDE& MR & \textbf{B-minFDE} \\
	\midrule
WIMP &3.09&7.71&0.84&1.47&2.90&0.42&- \\
Drivingfree&2.47&6.26&0.72&1.17&2.58&0.49&3.03\\
LGU&2.77&6.91&0.73&1.05&2.15&0.37&2.77\\
THOMAS & 1.95&4.71&0.64&0.88&1.51&0.20&2.16\\
QML* &1.84&4.98&0.62&\textbf{0.69}&1.39&0.19&1.95\\
BANet*&1.79&4.61&0.60&0.71&1.36&0.19&1.92\\
	\midrule
HeteroGCN&1.79&4.53&0.59&0.73&1.37&0.18&2.00\\
HeteroGCN-en*&\textbf{1.72}&\textbf{4.40}&\textbf{0.59}&\textbf{0.69}&\textbf{1.34}&\textbf{0.18}&\textbf{1.90} \\
	\bottomrule
\end{tabular}}
\vskip -0.1in 
\end{table*}

\subsection{Motion Prediction Decoder}
\label{sec,pred}
We adopt a goal-based decoder. It consists of three branches to predict $K$ future states and their respective scores.  All of these branches are designed as MLPs. 

The goal branch takes in the representation of agents output by GCMs $H_{P}^0$ and predicts $K$ goals $\{\bm{\hat{s}_T^{k}} \}_{k=1}^K$. Then, regression branch completes the trajectories $\hat{S}_f^{k}= \left[\bm{\hat{s}_{1}^{k}}, \bm{\hat{s}_{2}^{k}},...,\bm{\hat{s}_{T-1}^{k}} \right]$ conditioned on the embedding of  predicted goal $\bm{\hat{s}_{T}^{k}}$ and agent features $H_P^0$, $k=1, 2, \dots, K$. 
Finally, scoring branch estimates the confidence $\{\hat{q}^{k} \}_{k=1}^K$ of each prediction. 

\subsection{Learning}\label{sec.learn}
The loss function of the whole model consist of three parts, corresponding to goal prediction, trajectory completion, and score estimation, respectively.
\begin{equation}
    \mathcal{L}=  \lambda_1 \mathcal{L}_{goal} + \lambda_2  \mathcal{L}_{reg} + \lambda_3  \mathcal{L}_{score},
\end{equation}
with $\lambda_1, \lambda_2, \lambda_3$ as hyperparameters to balance the loss. Similar to previous methods \cite{liang2020learning,zeng2021lanercnn}, we choose the smooth $\ell_1$ loss  $l(\cdot)$ for $\mathcal{L}_{goal}$ and $\mathcal{L}_{reg}$, and the max-margin loss for $\mathcal{L}_{score}$. To obtain diverse predictions, $\mathcal{L}_{goal}$ and $\mathcal{L}_{reg}$ are computed based on the prediction achieving the minimum goal displacement error
\begin{equation}
       \mathcal{L}_{goal}= \frac{1}{|\mathcal{V}^0|} \sum_{v_i^0 \in \mathcal{V}^0}\min_{k}l(\bm{\hat{s}_{T}^{k},\bm{s_{T}} }) \label{e.goal_loss}
\end{equation}
\begin{equation}
     \mathcal{L}_{reg}= \frac{1}{|\mathcal{V}^0|(T-1)} \sum_{v_i^0 \in \mathcal{V}^0} \sum_{t=1}^{T-1} l(\bm{\hat{s}_{t}^{k^*},\bm{s_{t}}}),
\end{equation}
where $\{\bm{s_t}\}_{t=1}^T$ represent ground truth future states of agents and  $k^*$ indicates the index of the prediction achieving the minimum loss in Eq.~\eqref{e.goal_loss}.  The max-margin loss drives the model to assign the highest confidence score to the optimal prediction $\hat{S}_f^{k^*}$. Mathematically, with $\epsilon$ indicating the margin,
\begin{equation}
    \mathcal{L}_{score}= \frac{1}{|\mathcal{V}^0|(K-1)} \sum_{v_i^0 \in \mathcal{V}^0} \sum_{k \neq k^*} \max(\hat{q}^{k}-\hat{q}^{k^*} + \epsilon, 0).
\end{equation}

\section{Experimental Results}
We evaluate our model on the public Argoverse and Argoverse2 Motion Forecasting benchmarks, where the proposed model achieves significant improvements over a collection of state-of-the-art published methods. Furthermore, several ablation studies are conducted to verify the proposed modules.

\label{sec:result}
\subsection{Experimental Settings}

\begin{figure*}[tp]
\renewcommand{\baselinestretch}{1.0}
\renewcommand{\abovecaptionskip}{0pt}
\vskip 0.2in
\subfigure[A single model on straight roads.]{ 
\centering
  \includegraphics[width=0.32\linewidth]{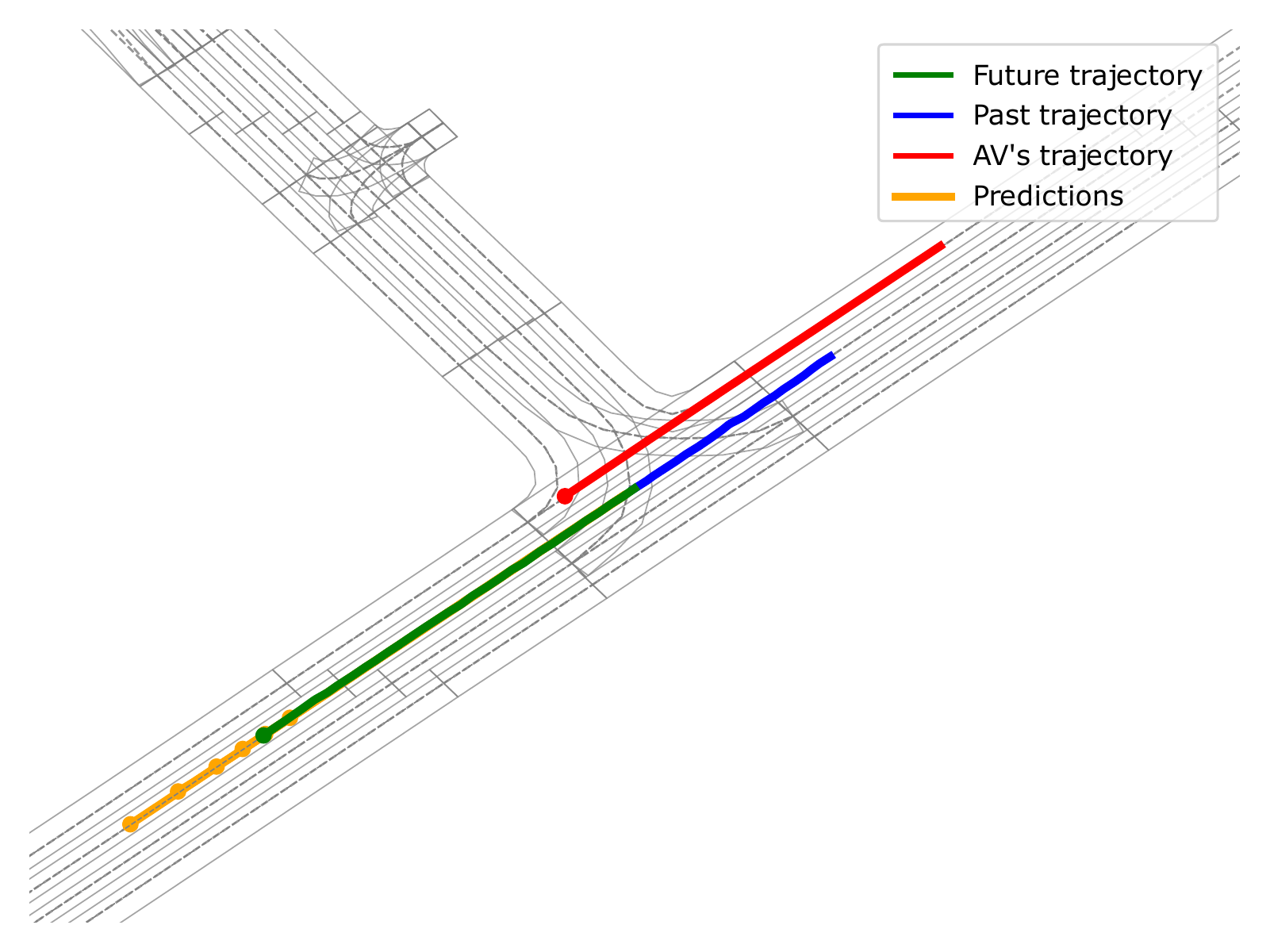}}
 \subfigure[A single model on intersections.]{ 
\centering
  \includegraphics[width=0.32\linewidth]{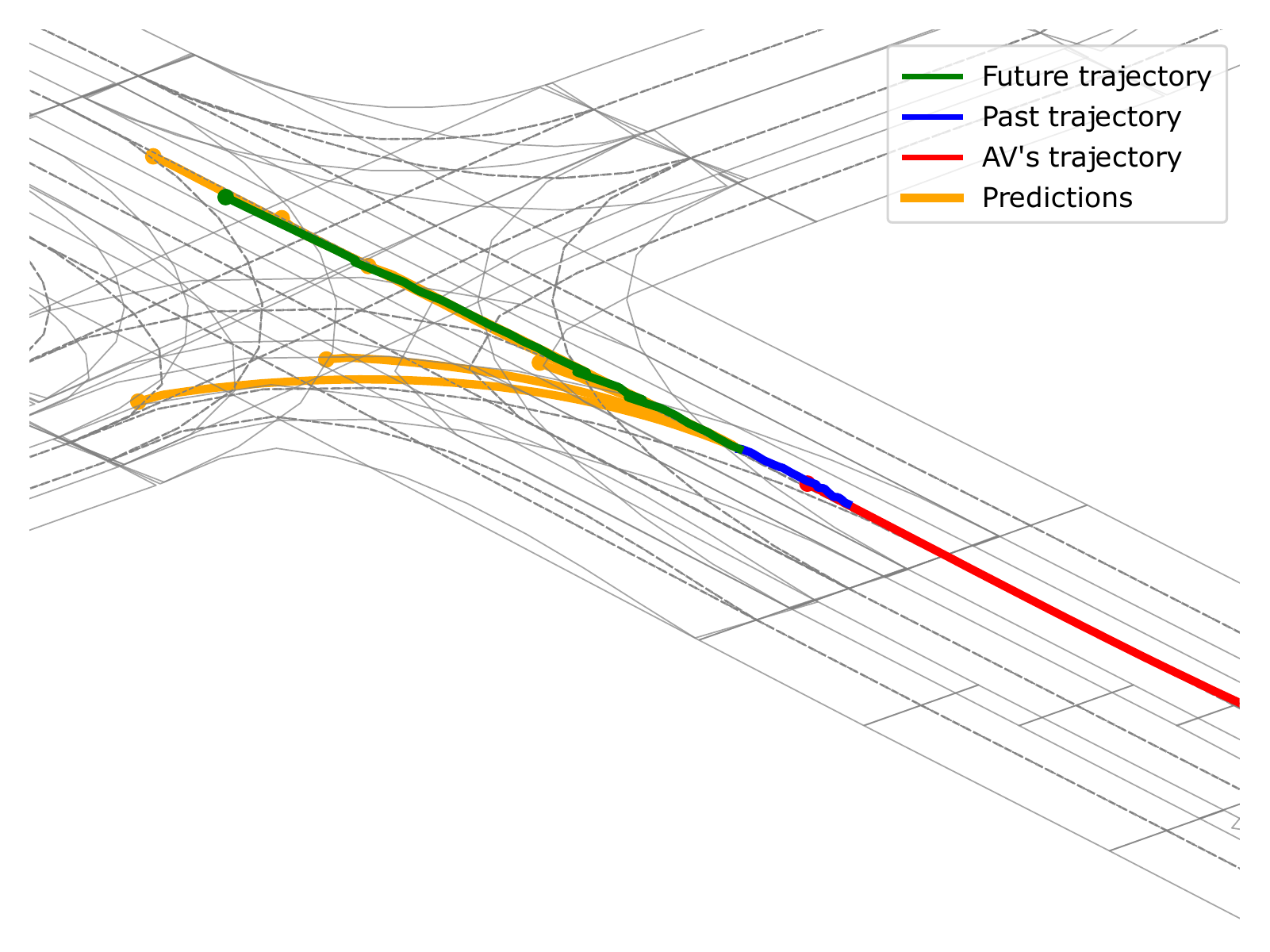}}
\subfigure[Self-ensemble results on intersections.]{ 
\centering
  \includegraphics[width=0.32\linewidth]{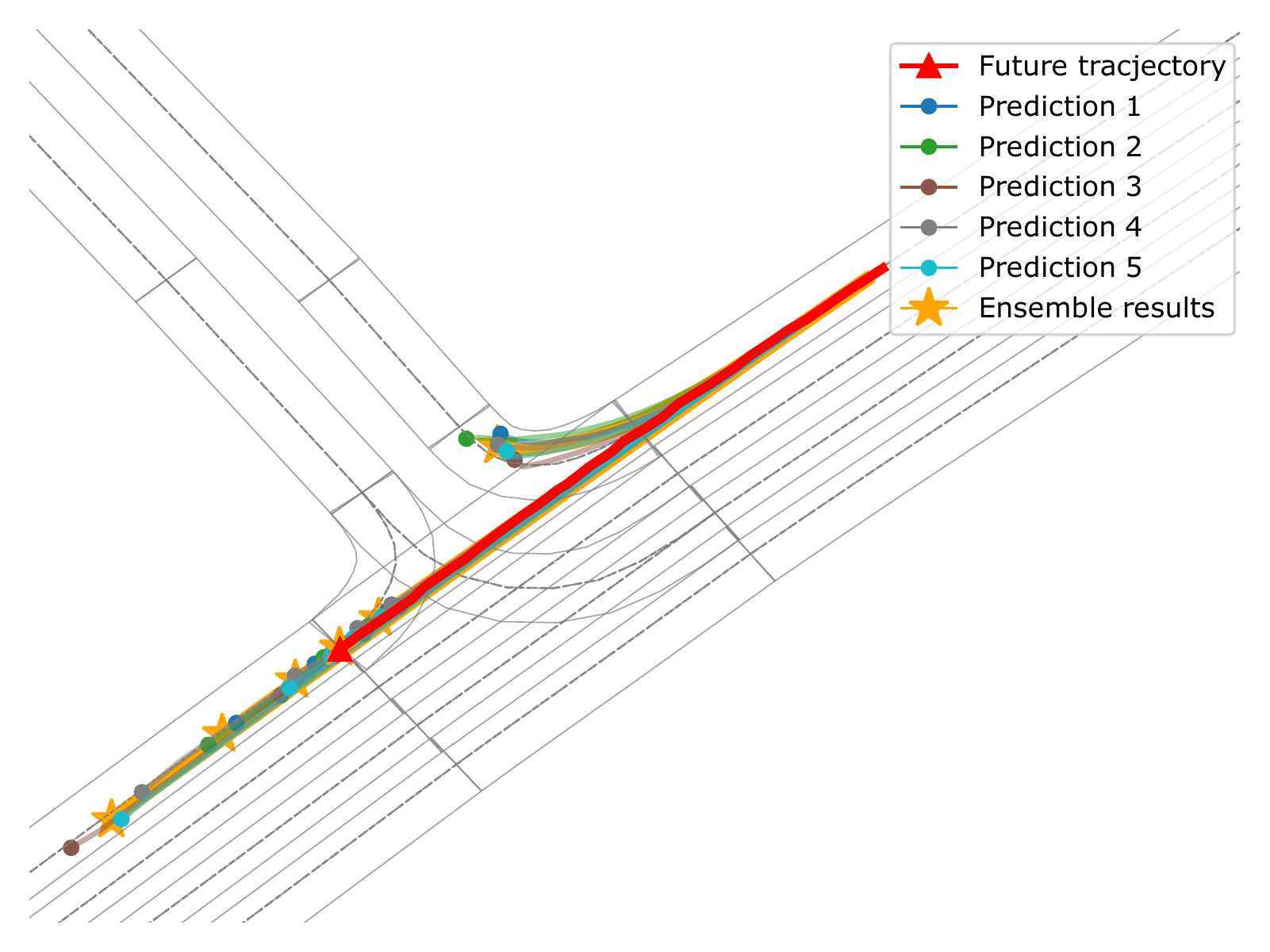}}
 \caption{Illustration of prediction results on the validation set of Argoverse. 
 For clarity, trajectories of surrounding vehicles are not presented and  self-ensemble is illustrated with five models. } \label{fig.2}
\vskip -0.05in
\end{figure*}

\begin{figure*}
\centering
\subfigure[Dense traffic.]{ 
\centering
  \includegraphics[width=0.49\linewidth]{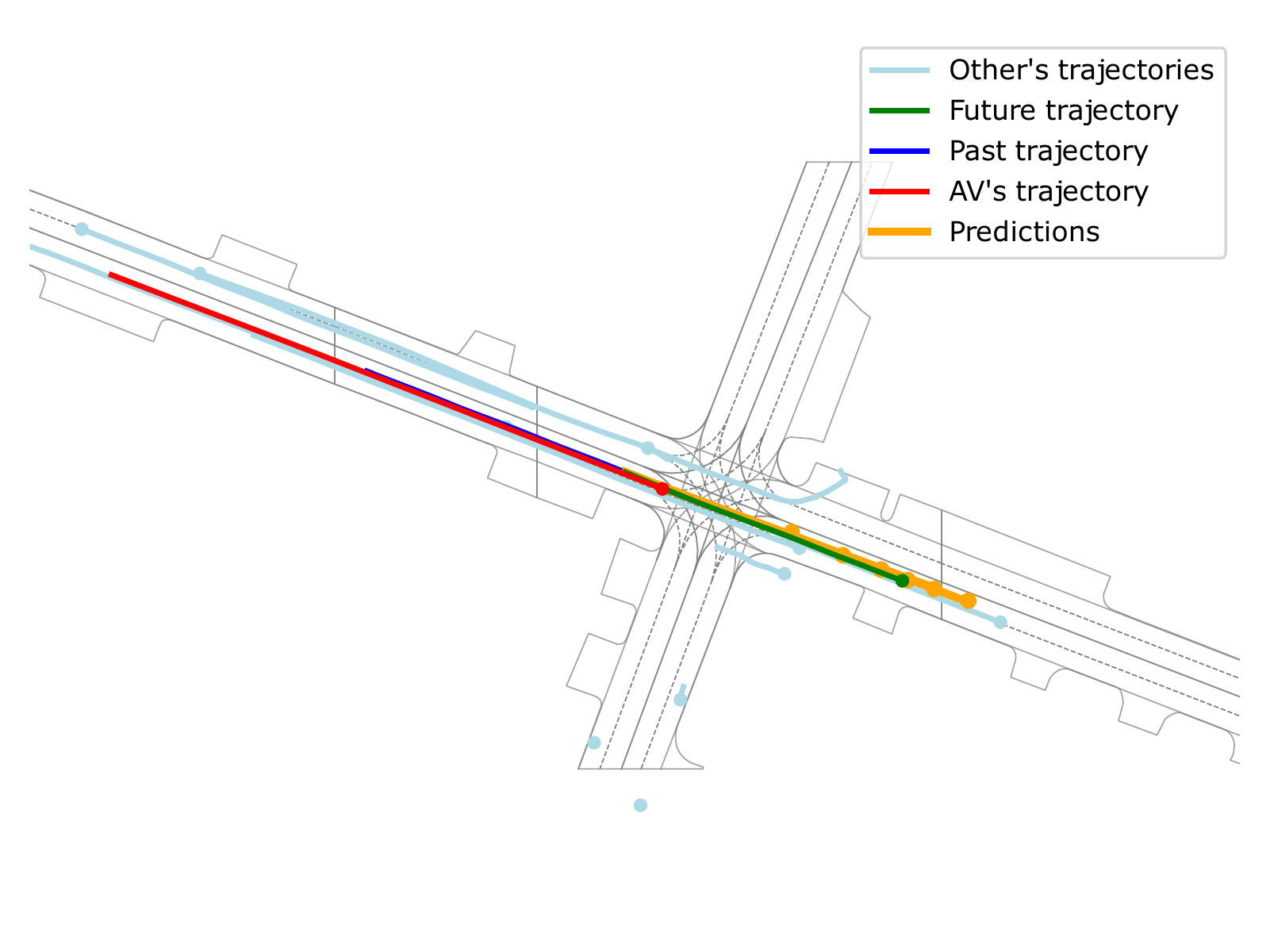}}
\hspace{0.25in}
 \subfigure[Intersections.]{ 
\centering
  \includegraphics[width=0.35\linewidth]{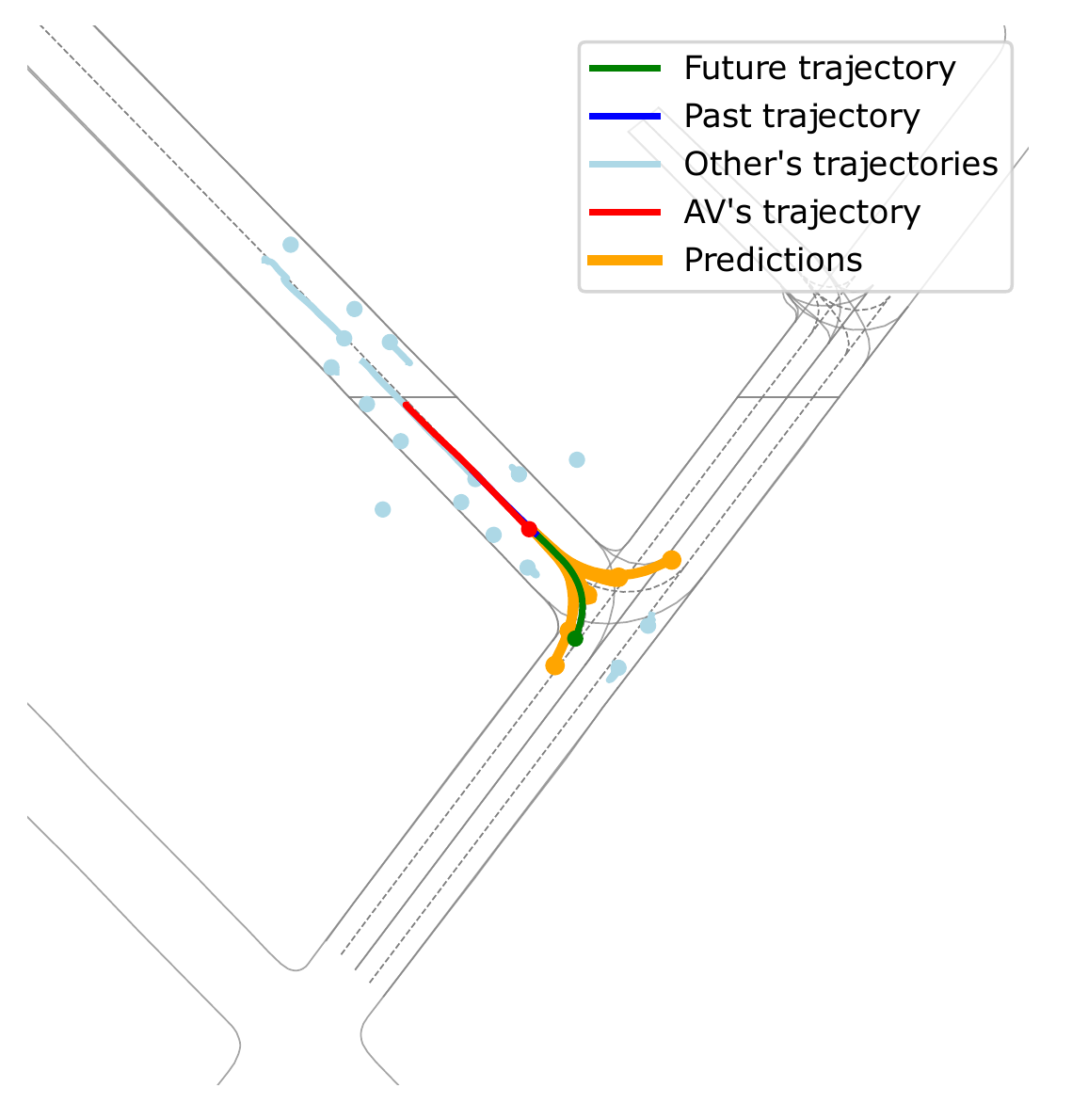}}
 \caption{Illustration of prediction results on the validation set of Argoverse2. } \label{fig.3}
\vskip -0.15in
\end{figure*} 

\textbf{Datesets.}  Argoverse \cite{chang2019argoverse} consists of 324,557 scenarios which are split into training, validation, and testing sets with the ratio $205,942:39,472:78,143$. The scenarios are generally selected from challenging cases, including interactions, turns, lane changes, and  dense traffic. Each scenario contains the locations of objects in 5 seconds  sampled at $10$ HZ, where initial 2 seconds (20 frames, $-T_{19}, \dots, T_0$) are adopted as observations to forecast the trajectory in the next 3 seconds (30 frames, $T_1, \dots, T_{30}$). A case typically consists of multiple moving agents, including an autonomous vehicle (AV) and other cars, and a single ``interesting'' agent is selected to predict per sequence. Locations of agents are provided as their states $\{\bm{s_t}\}$ and HD maps are also rendered.
 
Besides, Argoverse2 \cite{wilson2021argoverse} provides more challenging data with a longer forecast horizon, predicting 6 seconds based on 5 second observations, than Argoverse. It contains 250,000 scenarios with a split ratio of 8:1:1 for training, validation, and testing sets. Furthermore, compared with Argoverse, Argoverse2 contains 5 rather than 1 object types, including vehicle, pedestrian, cyclist, etc., each with dynamic and static categories. Agent states are further expanded into location, velocity, orientation, and one-hot encoding of agent types.

\textbf{Metrics.} According to the setting of Argoverse and Argoverse2 Motion Forecasting leaderboards, we predict $K$ trajectories for each ``interesting'' agent and report the following metrics: Minimum Average Displacement Error (minADE@K), Minimum Final Displacement Error (minFDE@K), Miss Rate (MR@K), Brier minimum Final Displacement Error (B-minFDE@K), with $K=1$ and $6$. ADE is the averaged $\ell_2$-norm distance between the prediction and the ground truth over all the time steps, and minADE@K refers to the minimum  of $K$ predictions.  Similarly, the minFDE@K represents the minimum $\ell_2$-norm distance between the prediction  of final position (goal) and the  ground truth. Based on minFDE@K, MR@K is defined as the ratio of scenarios where the minFDE@K is beyond 2 meters. Finally, the B-minFDE@K further considers the probability of the optimal prediction.

\textbf{Configurations.}
We adopt a scenario-wise coordinate system in the experiments, that is, all agents in the scenario share the same coordinate system.   We take the location of the agent to predict at $t=0$ as the origin, and adopt its current direction as $x$-axis. All the data are normalized accordingly. In graph construction, historical states $S_h= \left[ \bm{s_{-T'+1}},\bm{s_{-T'+2}},...,\bm{s_{0}} \right]$ are sliced into $P$ groups with interval $\tau=5$, $\emph{i.e.}$, $P=4$ in Argoverse and $P=10$ in Argoverse2. The $\psi(\cdot)$ and the $\nu(\cdot)$ in the graph convolution operator each use a linear layer followed by a ReLU activation function.
The GCM contains 2 and 3 graph convolution operators for Argoverse and Argoverse2, respectively. 
Models are trained with the Adam \cite{kingma2014adam} optimizer with batch size 128 for 40 and 90 epochs on Argoverse and Argoverse2.

\begin{figure*}[tp]
\renewcommand{\baselinestretch}{1.0}
\renewcommand{\abovecaptionskip}{0pt}
\vskip 0.2in
\subfigure[DenseTNT.]{ 
\centering
\includegraphics[width=0.33\linewidth]{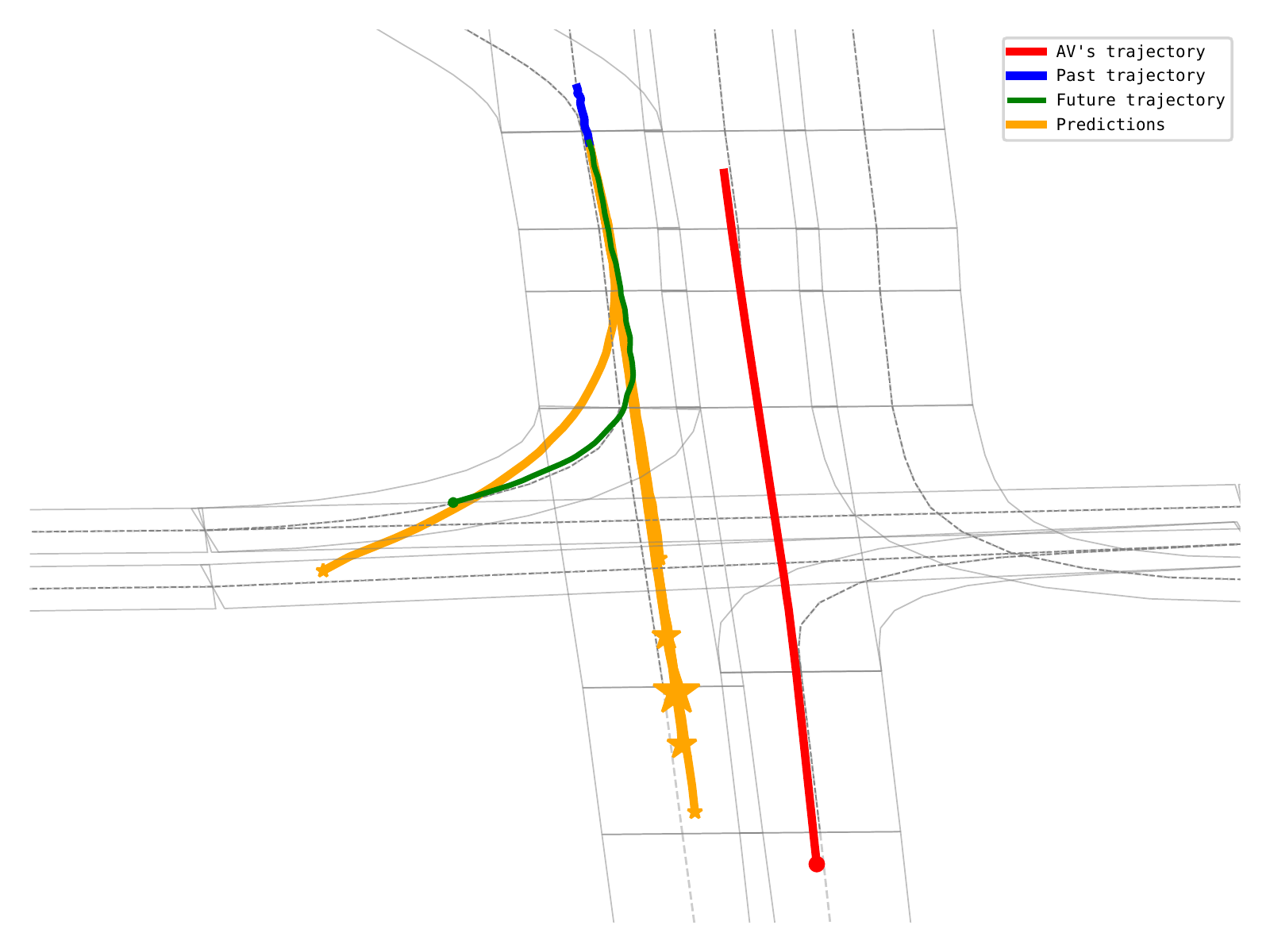}}
\subfigure[LaneGCN.]{ 
\centering
  \includegraphics[width=0.33\linewidth]{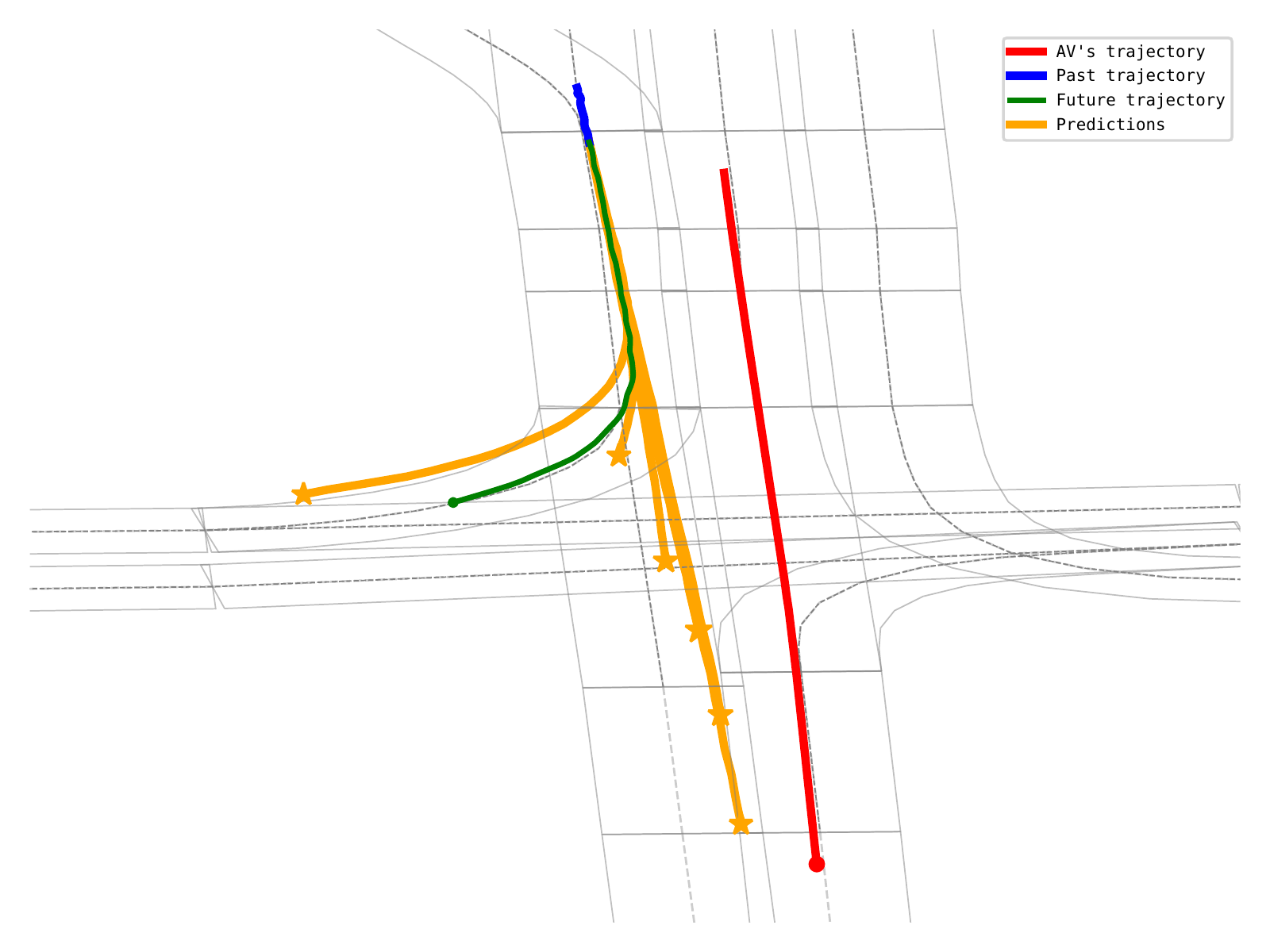}}
\subfigure[HeteroGCN.]{ 
\centering
  \includegraphics[width=0.33\linewidth]{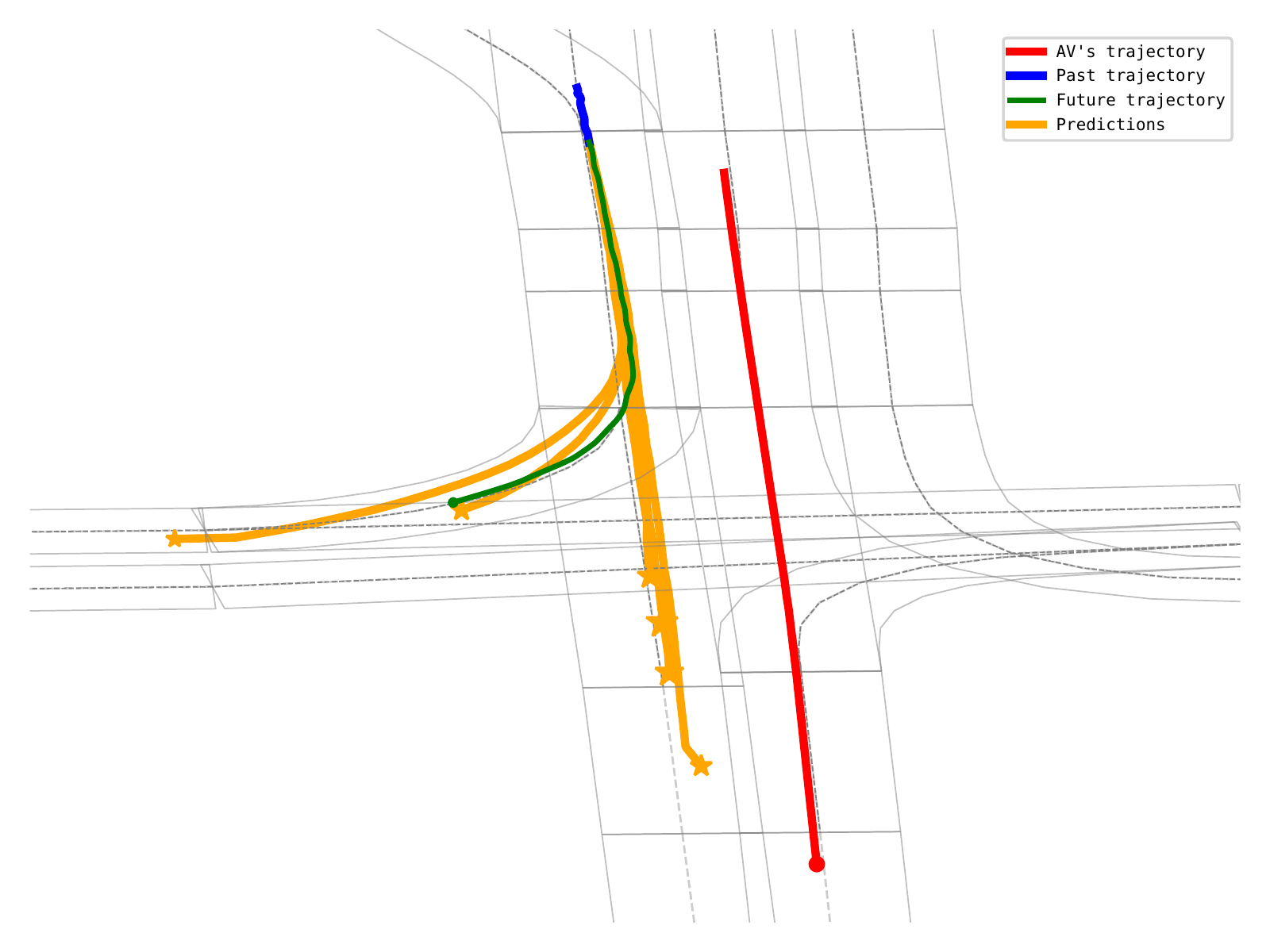}}
 \caption{Comparison of prediction results on the validation set of Argoverse using respective official code of baselines, with the goal of each predicted trajectory  represented by a pentagram and its size indicating the confidence.
 For clarity, trajectories of surrounding vehicles are not presented. } \label{fig.comp}
\end{figure*}

\textbf{Baselines.} We compare with a collection of state-of-the-art methods in motion forecasting.
First, \textbf{Prime} \cite{song2022learning} exploits a model-based generator and a learning-based evaluator to produce feasible future trajectories. Besides, \textbf{Jean} \cite{9197340} employs LSTMs and multi-head attention to handle trajectories of agents and their interactions. Then, a series of GNNs based methods are compared. 
\textbf{LaneGCN} \cite{liang2020learning} builds a lane graph for HD maps and captures various interactions through additional attention modules, and \textbf{BANet} \cite{DBLP:journals/corr/abs-2206-07934} as a variant further fuses lane boundary information.  Similarly, \textbf{Gohome} \cite{gilles2021gohome} and \textbf{THOMAS}\cite{gilles2022thomas} exploit a lanelet-level graph to encode HD map. 
\textbf{LaneRCNN} \cite{zeng2021lanercnn} further adopts agent-specific graphs and realizes interaction through pooling in the global lane graph. Besides,  social interactions are processed with graph-based attention in \textbf{WIMP} \cite{khandelwal2020if}. \textbf{DenseTNT} \cite{gu2021densetnt} relies on the scenario representation of Vectornet, a two-level graph, and generates trajectories from dense goal candidates. Furthermore, \textbf{SceneTransformer} \cite{ngiam2021scene} directly takes in sequence data and exploits transformers to handle interactions via factorized attention. Finally, \textbf{Multipath++} \cite{varadarajan2021multipath} proposes a context gating module to handle interactions and employs  model ensemble to enhance performance. 

\subsection{Results and Analysis}
As presented in Table~\ref{t:1} and Table~\ref{t.av2}, the proposed method outperforms all the baseline models in terms of B-minFDE@6, the official ranking metric on Argoverse and Argoverse 2. It still performs the best in accordance with most other metrics like minFDE.  
As presented in Fig.~\ref{fig.2} and Fig.~\ref{fig.3}, the predicted trajectories of HeteroGCN are realistic and diverse on both Argoverse and Argoverse2. HeteroGCN adopting dynamic graphs for scenario modeling achieves significant improvements over static road graph based methods, like LaneGCN, LaneRCNN, GoHome, and THOMAS. Compared with VectorNet and DenseTNT, the proposed model further takes into account the heterogeneity and evolution of nodes and edges, and effectively captures the topology of roads. In contrast with SceneTransformer, HeteroGCN explicitly encodes various interactions with graphs rather than implicitly through factorized attention, which facilitates the deep models to exploit high-order relationships between nodes. As shown in Fig.~\ref{fig.comp}, HeteroGCN produces more accurate predictions than baselines.

Following previous methods \cite{varadarajan2021multipath,DBLP:journals/corr/abs-2206-07934}, we further employ a self-ensemble strategy to enhance the robustness and accuracy of prediction. With a tiny  disturbance on the network architecture and hyper-parameters, we obtain eight submodels of the proposed method. Then the $k$-means algorithm is employed to cluster the produced states from these models into 6 categories.  The centers of the generated clusters are adopted as final predictions and their confidence scores are obtained by aggregating the initial scores of trajectories within respective clusters. As presented in Table~\ref{t:1} and Table~\ref{t.av2}, the proposed HeteroGCN-en performs the best among baselines with respective ensemble strategies, such as Multipath++ and BANet.

\subsection{Ablation Studies}
\textbf{Ablations on scenario modeling.} In order to evaluate the benefits of scenario modeling with dynamic heterogeneous graphs, we study the following variants.
\begin{itemize}
\item  \textbf{HomoGCN} ignores the heterogeneity in  nodes and edges in GCMs. It is achieved by taking a common message passing scheme and transformation for different $z$ and $r$ in Sec.~\ref{sec.gcn}.

\item  \textbf{HeteroGCN-static} employs a static graph $\mathcal{G}$, \textit{i.e.,} a common graph topology $A$ for all the graph convolution layers. Motion encoding module remains to processes all of the historical trajectories. 
The first GCM directly takes in the final hidden state from motion encoding module as agent node features (lane node features are the same as HeteroGCN), and following GCMs feed on the output of preceding GCMs to fuse road and agent information. Others are the same as HeteroGCN.

\item  \textbf{HomoGCN-static} further ignores the heterogeneity in nodes and edges in GCMs, compared with HeteroGCN-static. 
 \end{itemize}

\begin{table*}[tp]
\vskip 0.1in
\centering
 \caption{Ablation studies of scenario modeling on the testing set of Argoverse2. }\label{t.av2.mod}
\setlength{\tabcolsep}{1.5mm}{
\begin{tabular}{c|cc|ccc|ccc|c}
	\toprule
\multirow{2}{*}{Ablation variants}& \multirow{2}{*}{Heterogeneous}&\multirow{2}{*}{Dynamic}  &\multicolumn{3}{c|}{$K=1$}&\multicolumn{4}{c}{$K=6$}\\
& & & minADE & minFDE& MR & minADE & minFDE& MR & \textbf{B-minFDE} \\
\midrule
HomoGCN-static& & &1.90&4.87&0.61&0.77&1.46&0.20&2.10\\
HeteroGCN-static&\checkmark&&1.84&4.67&0.61&0.74&1.40&0.18&2.04 \\
HomoGCN&&\checkmark &1.85&4.69&0.60&0.74&1.39&0.18&2.04\\
HeteroGCN(proposed)&\checkmark&\checkmark&1.79&4.53&0.59&0.73&1.37&0.18&2.00\\
	\bottomrule
\end{tabular}}
\end{table*}
\begin{table*}[tp]
\centering
 \caption{Ablation studies of heterogeneous graph convolutional recurrent networks on the testing set of Argoverse2. }\label{t.av2.hgcn}
\setlength{\tabcolsep}{1.5mm}{
\begin{tabular}{c|ccc|c|ccc|ccc|c}
	\toprule
Motion 
 & \multicolumn{3}{c|}{GCM} &\multirow{2}{*}{Recurrent}&\multicolumn{3}{c|}{$K=1$}&\multicolumn{4}{c}{$K=6$}\\
Encoding& GCM-1&GCM-2&GCM-3&& minADE & minFDE& MR & minADE & minFDE& MR & \textbf{B-minFDE} \\
\midrule
\checkmark&& &&-&3.03&8.36&0.76&1.26&3.03&0.49&3.67\\
&\checkmark& &&\checkmark&2.02&5.05&0.63&0.80&1.50&0.20&2.14\\
\checkmark&\checkmark&&&\checkmark&1.88&4.79&0.61&0.75&1.43&0.19&2.07\\
\checkmark  &    &\checkmark& &\checkmark&1.84&4.68 &0.60 &0.74&1.39&0.19&2.03 \\ 
\checkmark& & & \checkmark&&1.85&4.71&0.61&0.75&1.41&0.19&2.06\\ 
\checkmark& & & \checkmark&\checkmark&1.79&4.53&0.59&0.73&1.37&0.18&2.00\\
	\bottomrule
\end{tabular}}
\end{table*}

As presented in Table~\ref{t.av2.mod},  compared with the baseline variant HomoGCN-static that  models scenarios as homogeneous static graphs and exploits a homogeneous version of the proposed GCNs, HeteroGCN-static and HomoGCN achieve better performance by further considering the heterogeneity and the evolution of scenario components and their interactions, respectively. Furthermore, considering the heterogeneity and evolution jointly, the proposed HeteroGCN performs the best.

\textbf{Ablations on the main components of HeteroGCNs.} As shown in Table~\ref{t.av2.hgcn}, all of the modules play a significant role in the scenario representation learning for motion forecasting. First, the motion encoding module that pre-encodes motion features of agents
increases the performance from 2.14 into 2.07 in terms of B-minFDE. Furthermore, the proposed graph convolution module (GCM) effectively handles multi-type interactions in  scenarios. Even with the GCM just composed of one graph convolution layer, the proposed model still outperforms all the single baselines presented in Table~\ref{t.av2}. With the increment of the number of graph convolution layers in GCM from~1~to~3, the performance keeps improving. Specially, designing the network as a recurrent one,  different snapshots of a dynamic graph are processed with a common GCM,  saves free-parameters and reduces the B-minFDE from 2.06 to 2.00.

\subsection{Computational Complexity and Running Time}
The computational complexity of HeteroGCNs is dominated by graph convolution with $O(|\mathcal{E}|d^2)$, $d$ indicating the dimension of node features and $\mathcal{E}=\bigcup_{r} \mathcal{E}^r$. In terms of running time, HeteroGCN is still competitive. The average inference time of LaneGCN, DenseTNT, and  HeteroGCN  is 55.32, 138.59, and 57.03 milliseconds on Argoverse with batch size 1,  with their official code on a computer with an RTX 2080 GPU, respectively.
           
\section{Conclusion}
\label{sec:conclusion}
In this paper, we have presented a framework to learn dynamic scenario representation and forecast future motions of agents in autonomous driving systems. First, the proposed driving scenario modeling strategy with dynamic graphs explicitly models complex interactions like agent-road and their evolution in the scenario. Furthermore, 
the designed heterogeneous graph convolutional recurrent network
learns to exploit multiple high-order interactions and spatio-temporal information in the scenario jointly yet differently and improves the motion forecasting performance on several challenging public benchmarks.  In future, it is interesting to extend the framework from motion forecasting to motion planning.

\bibliographystyle{IEEEtran}
\bibliography{IEEEabrv,example}
\end{document}